\documentclass[letterpaper]{article} %
\usepackage[preprint]{aaai2027} %
\usepackage[hyphens]{url}  %
\usepackage{graphicx} %
\usepackage{natbib}  %
\usepackage{caption} %
\usepackage{algorithm}

\usepackage[table]{xcolor} %
\usepackage{algpseudocode}
\usepackage{multirow}
\usepackage{amssymb}     %
\usepackage{array}

\usepackage[accsupp]{axessibility}  %

\usepackage{stackengine} %
\graphicspath{{images/}}
\usepackage{tabularx}

\newlength{\labw}\newlength{\imgw}
\usepackage{newfloat}
\usepackage{listings}
\DeclareCaptionStyle{ruled}{labelfont=normalfont,labelsep=colon,strut=off} %
\floatstyle{ruled}
\newfloat{listing}{tb}{lst}{}
\floatname{listing}{Listing}

\usepackage{booktabs}

\title{GateMOT: Q-Gated Attention for Dense Object Tracking}
\author{
    Mingjin Lv\textsuperscript{\rm 1},
    Zelin Liu\textsuperscript{\rm 1},
    Feifei Shao\textsuperscript{\rm 2},
    Yi-Ping Phoebe Chen\textsuperscript{\rm 3},\\
    Junqing Yu\textsuperscript{\rm 1},
    Wei Yang\textsuperscript{\rm 1},
    Zikai Song\textsuperscript{\rm 1}\corresponding
}
\affiliations{
    \textsuperscript{\rm 1}Huazhong University of Science and Technology, Wuhan, China\\
    \textsuperscript{\rm 2}Zhejiang University, Hangzhou, China\\
    \textsuperscript{\rm 3}La Trobe University, Melbourne, Australia\\
    \{lvmingjin77, skyesong\}@hust.edu.cn
}

\begin{document}

\maketitle

\begin{abstract}
    While large models demonstrate the strong representational power of vanilla attention, its quadratic all-to-all interactions are computationally prohibitive for dense motion estimation on high-resolution feature maps in dense object tracking. This mismatch limits the use of attention-based modeling for dense object tracking in crowded and occlusion-heavy scenes.
    To address this challenge, we introduce \textbf{GateMOT}, an online tracking framework centered on \textbf{Q-Gated Attention (Q-Attention)}, an efficient and spatially aware attention mechanism tailored to dense task-head routing rather than global token aggregation. Our key idea is to repurpose the Query from a similarity-conditioning term into a learnable gating unit. This Gating-Query (Gating-Q) produces a spatially adaptive gate that modulates Key features in an element-wise manner, enabling explicit relevance selection instead of costly global aggregation. Built on this mechanism, parallel Q-Attention heads transform one shared feature map into task-specific yet consistent representations for detection, motion estimation, and re-identification, yielding a tightly coupled multi-task decoder that removes quadratic pairwise spatial interactions. GateMOT achieves 48.4 HOTA, 67.8 MOTA and 64.5 IDF1 on BEE24-test, the best among compared methods, and demonstrates strong performance on additional dense object tracking benchmarks. These results show that Q-Attention is a simple, effective, and transferable building block for attention-based dense object tracking. Code is available at \url{https://github.com/lvmingjin77-ui/GateMOT}.
\end{abstract}


\section{Introduction}
\label{sec:intro}
Dense object tracking is a fundamental task in computer vision~\cite{song3,song7,ReTrack,OFFSET,REFINE}, supporting applications~\cite{song9,song12,song14,HABIT,INTENT,HINT} such as autonomous driving and robotics~\cite{guo2022review,pereira2022sort}. Its core challenge is to maintain reliable identities over time while jointly addressing detection, motion estimation, and re-identification. Although attention mechanisms~\cite{vaswani2017attention,song1,song2} have driven major advances in vision~\cite{Carion2020End,MELT,STABLE}, their use for dense motion estimation in dense object tracking remains limited~\cite{stone2021smurf}. Vanilla attention has quadratic complexity and is costly on the high-resolution feature maps used for dense prediction, especially in crowded and occlusion-heavy scenes, limiting its use in dense tracking heads~\cite{dao2024flashattention2}.

\begin{figure}[t]
  \centering
  \includegraphics[width=\linewidth,keepaspectratio]{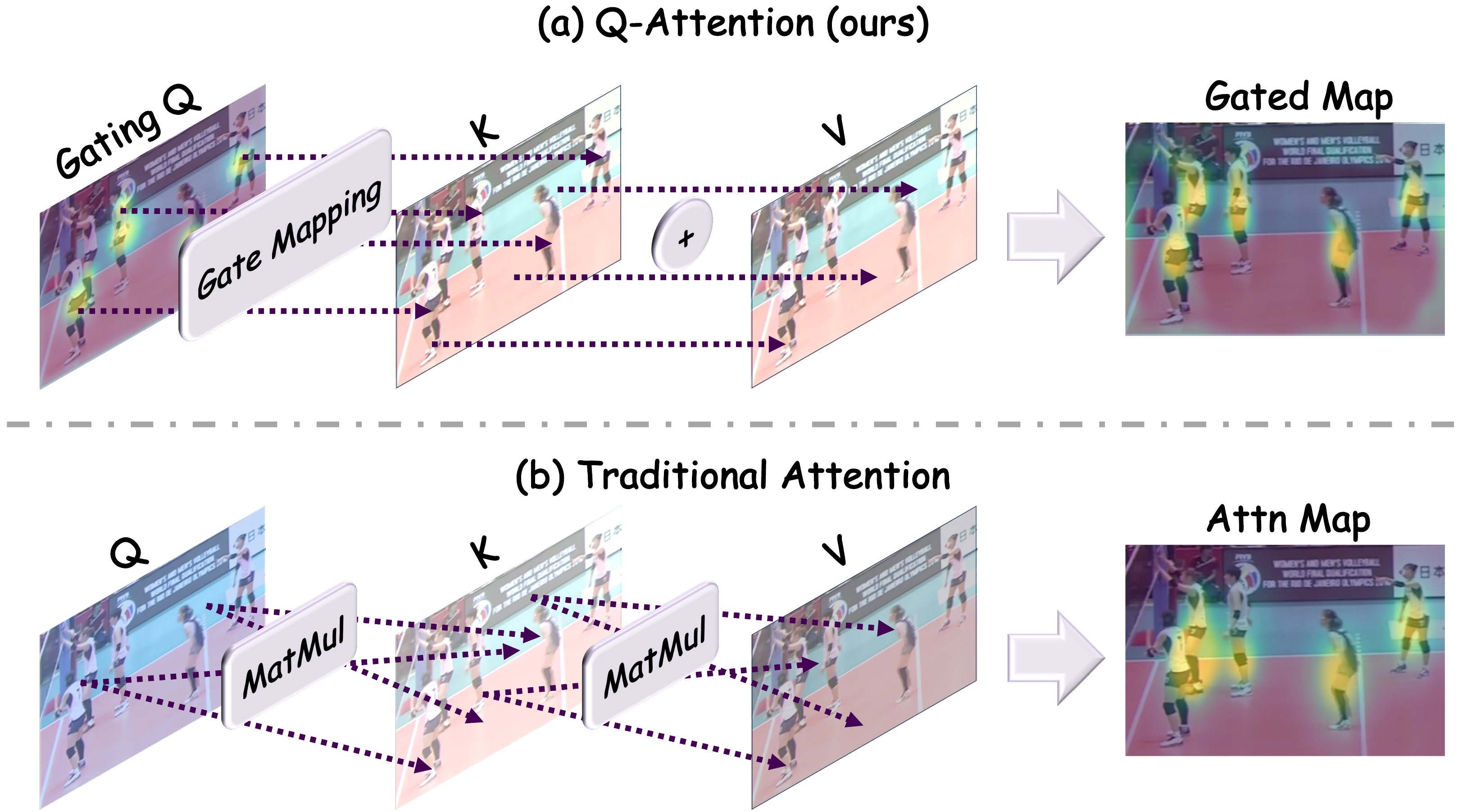} %
  \caption{
   \textbf{Comparison between our Q-Attention and vanilla attention.} \textbf{\textit{Top}}: our Q-Attention employs a learned Gating-Q to perform efficient, element-wise gating on Key (K) features. \textbf{\textit{Bottom}}: vanilla attention relies on two expensive matrix multiplications for global feature aggregation.
  }
  \label{fig:vs}
\end{figure}

This computational barrier is reflected in three representative design strategies: (1) tracking-by-detection methods with lightweight motion predictors~\cite{song4, song2025temporal}, where hand-crafted models such as the Kalman Filter~\cite{kalman1960new} keep motion estimation separate from deep visual features and can struggle with complex, non-linear dynamics~\cite{Cao2023Observation, Song_2026_CVPR}; (2) query-based association methods using sparse instance-level queries~\cite{meinhardt2022trackformer, zeng2022motr}, which restrict attention to a small set of queries to avoid dense computation but provide limited explicit geometric modeling of dense motion~\cite{Cai2022MeMOT}; and (3) deep motion learning approaches~\cite{wu2021track}, which infer dense motion from visual features but commonly rely on lightweight convolutional motion heads to maintain efficiency. Consequently, how to combine the feature-selection strength of attention with the efficiency required for dense motion estimation remains insufficiently explored.

To bridge this gap, we rethink attention for dense object tracking by transforming the role of Query. Instead of conditioning pairwise compatibility with Keys, Query is repurposed as a learnable \textbf{gating unit} that generates a dynamic, task-specific control signal for information flow~\cite{cheng2020refined}. This reformulation preserves Query-conditioned adaptive selection while redesigning it as efficient channel-spatial routing for dense prediction heads instead of global token aggregation.

Based on this principle, we build \textbf{GateMOT} within an established center-based online tracking paradigm. GateMOT is centered on \textbf{Q-Gated Attention (Q-Attention)}: as illustrated in Figure~\ref{fig:vs}, the \textbf{Gating-Query (Gating-Q)} learns a spatially adaptive gate that modulates Key features via efficient element-wise gating, replacing costly all-to-all interactions~\cite{dao2022flashattention}. The gated Key pathway is fused with an unfiltered Value pathway to preserve feature integrity. Because Q-Attention is lightweight enough to be used in every task head, GateMOT adopts a tightly coupled decoder that generates task-specific yet consistent features for detection, motion estimation, and re-identification while keeping computation efficient in dense settings.

Our contributions are summarized as follows:

\begin{itemize}
    \item We propose \textbf{Q-Gated Attention}, a compact dense task-head operator that transforms Query into a learnable channel-spatial gate for task-specific Key selection while retaining an unfiltered Value pathway, without pairwise token interaction.

    \item We instantiate Q-Attention in \textbf{GateMOT}, where parallel detection, motion-estimation, and re-identification heads derive task-specific yet spatially aligned representations from a shared dense feature map within an established center-based online tracking paradigm.

    \item Experiments across four benchmarks, controlled decoder and gating comparisons, and multiple encoder families demonstrate the accuracy--efficiency trade-off and transferability of Q-Attention.
\end{itemize}

\section{Related Work}
\label{sec:related_work}

\subsection{Kalman Filter-Based Methods}
The Tracking-by-Detection (TBD) paradigm~\cite{feichtenhofer2017detect} underlies many modern dense object tracking systems~\cite{zhang2022bytetrack, Cao2023Observation}. These methods decouple motion prediction from visual feature extraction for efficiency: state estimation is delegated to a Kalman Filter operating in a low-dimensional state space~\cite{kalman1960new}, while deep networks provide detection and, where applicable, appearance representations. Although computationally attractive, this separation often relies on a fixed linear motion model and prevents motion estimation from fully exploiting rich visual cues, which can hurt robustness under complex, non-linear dynamics~\cite{wojke2017simple, Cao2023Observation}. To address this limitation, GateMOT uses Q-Gated Attention (Q-Attention) to derive motion features directly from dense visual representations, reducing the reliance on fixed linear priors while keeping an efficient online tracking pipeline.

\subsection{Query-Based Methods}
A line of query-based dense object tracking methods, inspired by Transformer-based object detection~\cite{Carion2020End, zhu2021deformable} and multimodal reasoning~\cite{song15,song10,song11,song8,song5}, applies attention to a sparse set of instance-level track queries rather than dense feature maps~\cite{zeng2022motr, zhang2023motiontrack, yang2025motion}. Each query typically represents an object and attends across frames, thereby avoiding the cost of dense attention~\cite{gao2023memotr}. However, because attention primarily operates on sparse instance representations, motion is often modeled implicitly rather than predicted on dense feature maps, providing limited explicit geometric modeling of dense motion; persistent query--object binding can also be less flexible for frequent track births and deaths in dynamic scenes~\cite{Cai2022MeMOT}. To address this trade-off, GateMOT applies Q-Attention in dense heads to preserve explicit motion prediction and geometry-aware association without introducing quadratic pairwise spatial interactions.

\subsection{Learnable Motion-Based Methods}

Seeking to overcome the limitations of the Kalman Filter, another line of work develops learnable, data-driven models that infer motion or temporal displacement from visual features~\cite{tokmakov2021learning, qin2023motiontrack, Song_2026_CVPR}. While more flexible than rigid hand-crafted models, these approaches commonly use lightweight convolutional motion heads to remain tractable; motion and detection may share base features, but task-specific feature interaction within dense heads remains limited~\cite{ egu2024walker, Cai2022MeMOT}. To address this gap, GateMOT adopts Q-Attention as a lightweight attention primitive, enabling efficient feature interaction and task-specific specialization across detection, motion, and ReID in a unified decoder.

\section{Challenge and Motivation}
\label{sec:problem}

\paragraph{Structural Challenge in Dense Object Tracking.}
Dense object tracking systems generally use decoupled or coupled architectures. Decoupled pipelines optimize detection, motion, and association separately, yielding stable runtime behavior but weak conditioning of motion and identity reasoning on dense visual evidence under occlusion. Coupled pipelines share representations and better preserve geometric and temporal consistency, but task-specific feature interaction becomes costly on high-resolution feature maps. This creates a structural conflict: dense decoders need task-specific interaction precisely where computation is most constrained. Although object centers are sparse, dense decoding remains necessary to preserve spatial consistency and jointly optimize detection, motion estimation, and re-identification over a shared feature map.

Let $\mathbf{F}_t\in\mathbb{R}^{B\times C\times H_f\times W_f}$ be the shared dense feature map for frame $t$, where $B$ is the batch size and $H_f\times W_f$ is the feature resolution. The decoder must produce task-specific features
\[
\left(\mathbf{F}_t^{det},\ \mathbf{F}_t^{mot},\ \mathbf{F}_t^{id}\right) = \mathcal{A}(\mathbf{F}_t),
\]
for detection, motion estimation, and re-identification. We focus on this coupled parallel decoding setting because dense tracking quality depends on synchronized signals: accurate localization anchors motion cues, motion prediction stabilizes temporal association, and identity features reduce ambiguity under overlap. When these signals are loosely coupled, stage-wise errors can accumulate in crowded sequences.

\paragraph{Limitations of Attention and Convolution in Dense Decoding.}
Let $N=H_f\times W_f$ denote the number of spatial locations and $d$ the channel width. Vanilla self-attention~\cite{vaswani2017attention} incurs an $O(N^2d)$ pairwise interaction term in addition to projection cost, making it expensive for high-resolution dense heads. Linear attention~\cite{katharopoulos2020transformers,choromanski2021rethinking} reduces the interaction complexity to $O(Nd^2)$, but still performs global aggregation, which may mix irrelevant long-range responses with local instance evidence. Deformable attention~\cite{zhu2021deformable} reduces pairwise cost through sparse sampling, but still requires learned sampling and aggregation across multiple dense task heads. Thus, attention-based dense decoding faces a trade-off between expensive all-pairs interaction and reduced-complexity mechanisms that retain global or sparse aggregation overhead.

Standard convolutional heads are efficient and locality-preserving, but their input-independent kernels lack explicit Query-conditioned routing for task-specific feature selection. Under dense overlap, this can limit their ability to emphasize motion- or identity-sensitive cues at each location, reducing task specialization in coupled parallel decoding.

\paragraph{Problem objective.}
Our objective is to design $\mathcal{A}$ to retain Query-conditioned, task-specific feature selection in dense parallel heads without pairwise spatial interaction. GateMOT addresses this objective through Q-Attention, a gated aggregation operator that replaces token-pair weighting with pointwise Query-guided Key filtering and local aggregation. An ungated Value pathway preserves information independently of this selective route. Removing the $O(N^2d)$ interaction term yields $O(Nd^2)$ overall complexity and an $O(Nd)$ gating interaction. Its spatial cost therefore scales linearly with $N$, keeping dense multi-task decoding practical. Q-Attention thus preserves attention-style selectivity and produces task-specific, spatially aligned representations for detection, motion estimation, and re-identification.

\section{Method}
\label{sec:method}
This section presents \textbf{GateMOT} and its core mechanism, \textbf{Q-Gated Attention (Q-Attention)}. We first formulate Q-Attention, then describe its integration into GateMOT for detection, motion estimation, and re-identification, followed by extensions across encoder families.

\begin{figure}[t]
  \centering
  \includegraphics[width=\columnwidth]{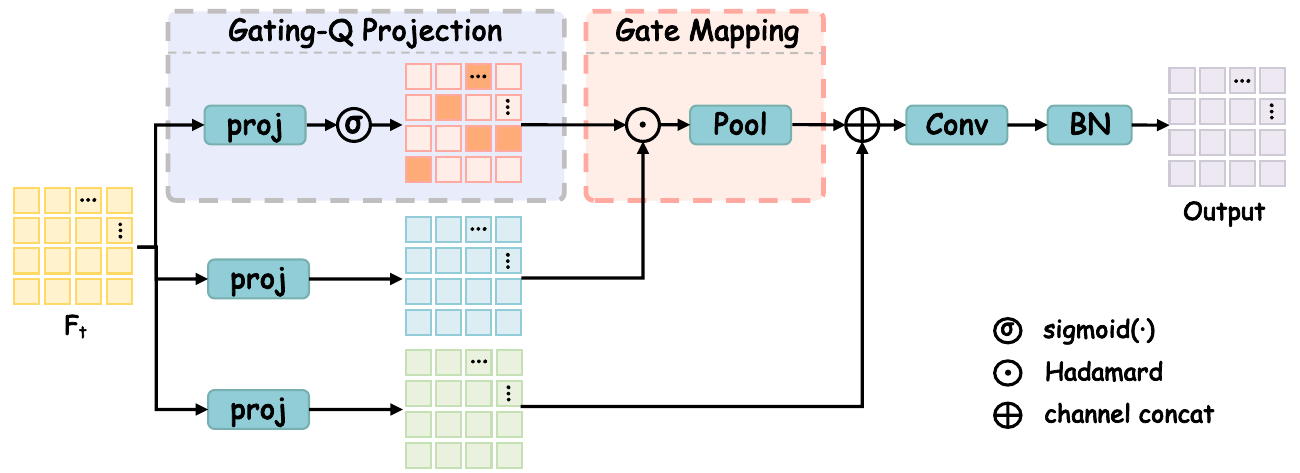}
  \caption{\textbf{The architecture of Q-Attention.} Gating-Q generates a channel-spatial relevance gate that selectively modulates Key features without constructing pairwise affinities. The selected Key evidence is locally aggregated and fused with an ungated Value residual pathway, balancing task-specific selection with information preservation.}
  \label{fig:q_attention_arch}
\end{figure}

\begin{figure*}[t]
  \centering
  \includegraphics[width=\textwidth]{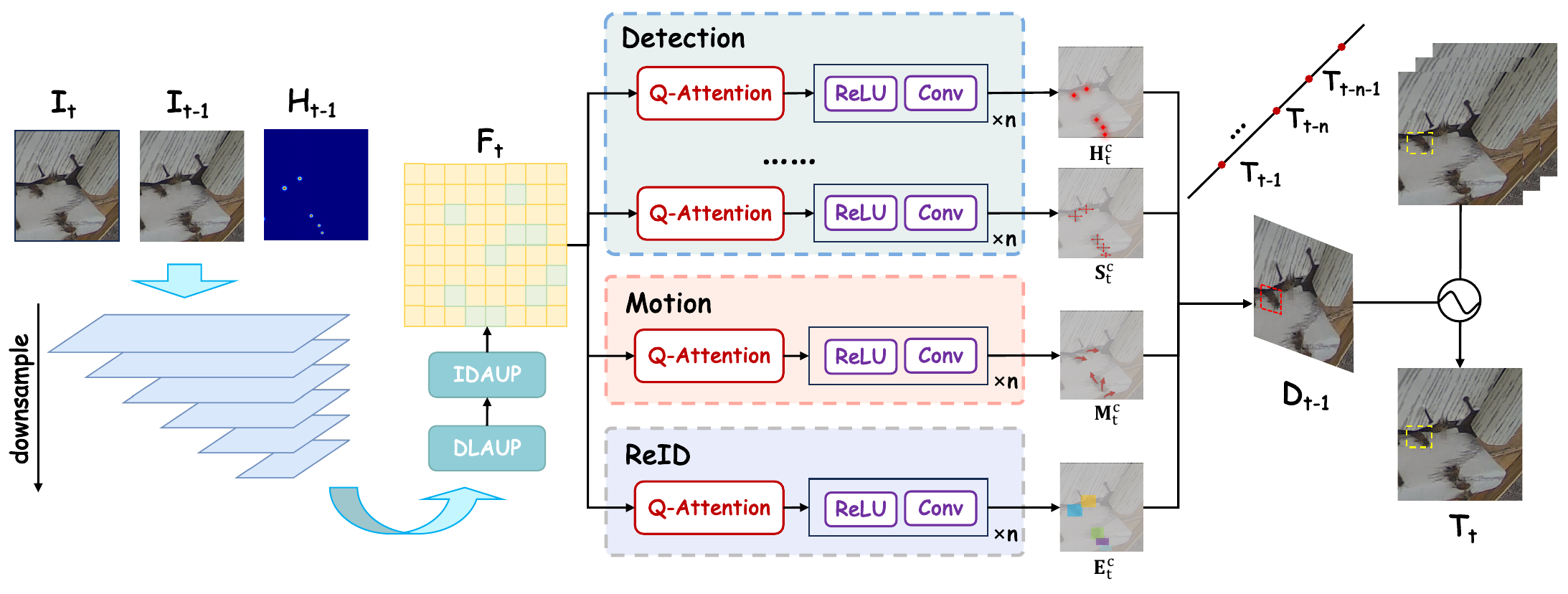}
  \caption{\textbf{The overall architecture of GateMOT.} Given the current frame $I_t$, the previous frame $I_{t-1}$, and the previous center heatmap $\mathbf{H}_{t-1}^{\mathrm{c}}$, the encoder produces a high-resolution feature map $\mathbf{F}_t$. Multiple Q-Attention prediction heads, grouped into detection, motion estimation, and ReID, generate a center heatmap, center and box offsets, motion vectors, ReID embeddings, and an auxiliary box-size map. Motion vectors back-project current detections to frame $t\!-\!1$ for matching with existing trajectories $\mathcal{T}_{t-1}$, producing updated tracks $\mathcal{T}_t$.}
  \label{fig:overall}
\end{figure*}

\subsection{Q-Gated Attention}
\label{subsec:qgla}

\paragraph{Mechanism.}
Given a shared feature map $\mathbf{F}\in\mathbb{R}^{B\times C\times H_f\times W_f}$, three independent $1\times1$ projections produce
\begin{equation}
\mathbf{Q}=W_Q(\mathbf{F}),\qquad
\mathbf{K}=W_K(\mathbf{F}),\qquad
\mathbf{V}=W_V(\mathbf{F}),
\end{equation}
where $\mathbf{Q},\mathbf{K},\mathbf{V}\in\mathbb{R}^{B\times C'\times H_f\times W_f}$, $B$ is the batch size, $H_f\times W_f$ is the feature resolution, and $C'$ is the head channel dimension. Conventional attention uses Query to construct pairwise compatibility with Keys and aggregate Values globally. Q-Attention instead converts Query into an input-dependent controller of information flow, replacing affinity aggregation with explicit selection and residual preservation. It is neither an approximation nor a drop-in replacement for global self-attention; its purpose is to retain Query-conditioned adaptive selection while avoiding pairwise spatial interaction in dense task heads. The three projections are functionally asymmetric: Query estimates task relevance, Key provides candidate evidence to be selectively routed, and Value forms a residual pathway independent of the gate.

As illustrated in Figure~\ref{fig:q_attention_arch}, Gating-Q produces a probabilistic channel-spatial relevance gate through a sigmoid function:
\begin{equation}
\mathbf{G} = \sigma(\mathbf{Q}).
\end{equation}
Each coefficient in $\mathbf{G}\in(0,1)^{B\times C'\times H_f\times W_f}$ independently controls the information flow at one channel-location pair. Unlike spatial Softmax, this independent gating permits multiple relevant responses without competitive normalization. Importantly, Q-Attention does not construct pairwise Query-Key affinities; Query directly modulates Key evidence at the same channel and spatial location.

The gate performs task-conditioned evidence routing through an efficient element-wise interaction:
\begin{equation}
\mathbf{K}' = \mathbf{G} \odot \mathbf{K}.
\end{equation}

This operation converts the shared representation into a selectively filtered Key stream without forming an $(H_fW_f)\times(H_fW_f)$ affinity matrix. The gated Key is then locally consolidated by
\begin{equation}
\bar{\mathbf{K}}=\mathrm{MaxPool}_{3\times3,\,s=1,\,p=1}(\mathbf{K}').
\end{equation}

Local max aggregation retains the strongest task-relevant response in each neighborhood, matching the peak-oriented structure of center-based dense prediction. In parallel, Value bypasses the selective gate and forms an ungated residual pathway, protecting fine-grained information from irreversible suppression by the gated Key stream. The two pathways are fused as
\begin{equation}
\mathbf{Y}=\mathrm{BN}\!\left(
\mathrm{Conv}_{1\times1}\!\left([\mathbf{V};\bar{\mathbf{K}}]\right)
\right),
\end{equation}
where $[\cdot;\cdot]$ denotes channel-wise concatenation. Although fusion is implemented by concatenation rather than direct addition, the Value branch serves a residual role by bypassing the selective information bottleneck. The gated Key pathway contributes task-selective evidence, whereas the Value residual pathway preserves ungated detail; their adaptive fusion balances specialization and information completeness. Because every prediction head learns independent $W_Q$, $W_K$, and $W_V$ projections from the same shared feature map, Q-Attention produces task-specific representations while maintaining a common spatial coordinate system.

\paragraph{Complexity.}
Let $N=H_f\times W_f$. The projections require $O(NCC')$, element-wise gating and local aggregation require $O(NC')$, and fusion requires $O(NC'^2)$. When $C$ and $C'$ are represented by a common channel width $d$, Q-Attention has $O(Nd^2)$ overall complexity with an $O(Nd)$ gating-and-aggregation term. In contrast, self-attention introduces an additional $O(N^2d)$ pairwise interaction term. Removing token-pair construction changes the dominant spatial dependence from quadratic to linear, allowing the selection--preservation mechanism to be instantiated independently in every dense prediction head.

\subsection{GateMOT for Tracking}
\label{subsec:gatemot}

\paragraph{Overall Architecture.}
Figure~\ref{fig:overall} illustrates GateMOT, which adopts established temporal inputs, center-based decoding, and motion backtracking~\cite{zhou2020tracking}, while introducing Q-Attention for task-specific dense decoding. Keeping this tracking scaffold fixed isolates the contribution of the decoder. Given a video sequence $\{I_t\}_{t=1}^{T}$, the tracker maintains trajectories $\mathcal{T}_t=\{\tau_i\}$, each carrying a bounding box $\mathbf{b}_i^t$ and appearance embedding $\mathbf{e}_i^t$. A DLA encoder, with DLA-34 as default, maps $I_t$, $I_{t-1}$, and $\mathbf{H}_{t-1}^{\mathrm{c}}$ to $\mathbf{F}_t=\phi(I_{t-1},I_t,\mathbf{H}_{t-1}^{\mathrm{c}})$.

From $\mathbf{F}_t$, parallel Q-Attention heads for detection, motion estimation, and re-identification predict the center heatmap $\mathbf{H}_t^{\mathrm{c}}$, center offsets $\mathbf{O}_t$, amodal box offsets $\mathbf{L}_t$, motion vectors $\mathbf{M}_t$, ReID embeddings $\mathbf{E}_t$, and auxiliary box-size map $\mathbf{S}_t$. Peaks on $\mathbf{H}_t^{\mathrm{c}}$ form detection candidates
\[
\mathcal{D}_t=\{(\mathbf{b}_k^{t},s_k^{t},
\mathbf{e}_k^{t},\mathbf{m}_k^{t})\}_{k=1}^{N_t},
\]
where $N_t$ is the number of candidates. The box $\mathbf{b}_k^t$ is decoded from $\mathbf{H}_t^{\mathrm{c}}$, $\mathbf{O}_t$, and $\mathbf{L}_t$, while $\mathbf{e}_k^t$ and the forward displacement $\mathbf{m}_k^t=\mathbf{c}_k^t-\mathbf{c}_k^{t-1}$ are sampled from $\mathbf{E}_t$ and $\mathbf{M}_t$ at center $\mathbf{c}_k^t$.

For online association, each center is back-projected to frame $t\!-\!1$ by
\[
\tilde{\mathbf{c}}_k^{t-1}
=\mathbf{c}_k^{t}-\mathbf{m}_k^{t}.
\]
The corresponding box is translated to $\tilde{\mathbf{c}}_k^{t-1}$ and matched to active trajectories by a first-stage Hungarian assignment using motion-compensated geometry. Remaining detections are matched to lost trajectories by cosine similarity on ReID embeddings under geometric constraints. Matched appearance templates use a confidence-adaptive update that favors reliable observations and suppresses noisy updates under occlusion:
\begin{equation}
\begin{aligned}
\tilde{\mathbf{e}}_i^{t}
&=\gamma(1-s_k^t)\mathbf{e}_i^{t-1}
+(1-\gamma)s_k^t\mathbf{e}_k^t,\\
\mathbf{e}_i^t
&=\frac{\tilde{\mathbf{e}}_i^t}
{\lVert\tilde{\mathbf{e}}_i^t\rVert_2}.
\end{aligned}
\end{equation}
Here, $\mathbf{e}_i^{t-1}$ is the stored template, $\mathbf{e}_k^t$ is the current embedding, $s_k^t\in[0,1]$ is the detection confidence, and $\gamma\in(0,1)$ is the smoothing factor. The updated template is $\ell_2$-normalized before storage. Unmatched high-confidence detections initialize new trajectories.

\paragraph{Gated Attention Decoder.}
The decoder follows a shared-to-specialized principle. A common $\mathbf{F}_t$ preserves spatial alignment, while independent Q-Attention heads construct task-specific information routes. Each head contains one Q-Attention block followed by a task-specific output projection and optional convolutional refinement. Because Q-Attention scales linearly with spatial locations, it can be deployed in every head without pairwise interaction. Independent gates adapt shared evidence to each prediction objective while retaining a common dense coordinate system. This design reduces interference among localization-sensitive, motion-sensitive, and identity-sensitive cues without separating the tasks from their shared spatial foundation.

\subsection{Extensions of Q-Attention}
\label{subsec:extension}

Q-Attention operates on a generic dense feature tensor and preserves its spatial resolution, allowing it to serve as a plug-compatible task-specialization layer across different encoder families. Migration only replaces the task-head feature selection block while keeping the dense object tracking protocol and supervision settings unchanged.

This controlled transfer isolates the effect of Q-Attention and shows that the selection--preservation principle is not tied to a single encoder architecture. Cross-backbone evidence is reported in the ablation studies.

\section{Experiments}
\label{sec:experiments}
 
\subsection{Settings}
 
\paragraph{Datasets.}
We evaluate GateMOT on BEE24~\cite{Cao2025TOPIC}, SportsMOT~\cite{Cui2023Sportsmot}, MOT17~\cite{dendorfer2021motchallenge}, and MOT20~\cite{Dendorfer2020Mot20}. We train separate models without external tracking data. BEE24 has no official validation split: we train on the first half of each training sequence and select checkpoints, designs, and thresholds on the second half without test feedback. After selection, we retrain on full train for final BEE24-test evaluation. Following TOPICTrack~\cite{Cao2025TOPIC}, BEE24 ablations use the test set; MOT17 ablations use MOT17-val. MOT17 and MOT20 use private detections from our center head.
 
\paragraph{Evaluation Metrics.}
We use HOTA~\cite{luiten2021hota} as the primary metric, which jointly measures detection accuracy (DetA) and association accuracy (AssA). We also report MOTA, which summarizes false positives (FP), false negatives (FN), and identity switches, and IDF1 for temporal identity consistency. For BEE24-test, we additionally report AssA and the number of identity switches (IDs), whereas MOT17, MOT20, and SportsMOT report AssA and DetA.
 
\subsection{Implementation Details}
 
\paragraph{Architecture.}
GateMOT uses ImageNet-pretrained DLA encoders~\cite{yu2018deep}, adopting DLA-34 by default and DLA-60 for SportsMOT. Its Gated Attention Decoder applies parallel Q-Attention heads to multi-scale features for detection, motion estimation, and re-identification. These heads predict center heatmaps and offsets, amodal boxes, an auxiliary box-size map, short-term motion vectors, and ReID embeddings.
 
\paragraph{Training Details.}
The framework is trained end-to-end with a multi-task objective: Focal Loss for center heatmap prediction, Cross-Entropy Loss for ReID classification, L1 Loss for center-offset, auxiliary box-size, and motion regression, and L1 and SIoU~\cite{gevorgyan2022siou} losses for amodal box regression. Optimization uses Adam with dataset-specific schedules: MOT17 is trained for 70 epochs (batch size 16, initial learning rate $1.25\times10^{-4}$, decayed at epoch 60), and BEE24 for 60 epochs (batch size 24, initial learning rate $2\times10^{-4}$, decayed at epoch 40). MOT20 and SportsMOT follow dataset-specific schedules.
 
\paragraph{Inference Details.}
During inference, we use a motion-guided backtracking matcher. Only detections with confidence above 0.4 are used for association.
 
\subsection{Benchmark Results}
We compare GateMOT with representative state-of-the-art methods on BEE24-test, SportsMOT, MOT17, and MOT20.
 
\paragraph{Results on BEE24-test.}
BEE24 contains extremely small, densely packed, and visually similar objects under frequent occlusion. GateMOT achieves the best HOTA (48.4), IDF1 (64.5), and the fewest identity switches (1,058) among the compared methods (Table~\ref{tab:gate_bee24}), indicating that the selection--preservation design maintains stable identities under severe occlusion and appearance ambiguity.
 
\begin{table}[t]
  \centering
    \caption{Comparison with representative methods on \textbf{BEE24-test}.}
    \label{tab:gate_bee24}
    \resizebox{\columnwidth}{!}{%
      \setlength{\tabcolsep}{4pt}
      \begin{tabular}{@{}l @{\hspace{4pt}\vrule\hspace{4pt}} ccccc@{}}
        \toprule
        \textbf{Tracker} & \textbf{HOTA} & \textbf{MOTA} & \textbf{IDF1} & \textbf{AssA} & \textbf{IDs}$\downarrow$ \\
        \midrule
 
        \textit{Kalman filter:} \\
        UniTrack~\cite{wang2021do}      & 41.6 & 54.6 & 53.0 & 34.8 & 1,972 \\
        FairMOT~\cite{zhang2021fairmot}        & 42.3 & 40.9 & 54.3 & 42.5 & 3,968 \\
        OC\text{-}SORT~\cite{Cao2023Observation} & 42.7 & 61.6 & 55.3 & 36.8 & 1,435 \\
        ByteTrack~\cite{zhang2022bytetrack}    & 43.2 & 59.2 & 56.8 & 38.3 & 1,303 \\
        TOPICTrack~\cite{Cao2025TOPIC}          & 46.6 & 66.7 & 59.7 & 40.3 & 1,401 \\
        \midrule
 
        \textit{Query\text{-}based:} \\
        TrackFormer~\cite{meinhardt2022trackformer} & 44.3 & 41.5 & 53.9 & 42.3 & 3,405 \\
        \midrule
 
        \textit{Learnable motion:} \\
        TraDeS~\cite{wu2021track}       & 30.9 & 42.2 & 34.8 & 20.2 & 5,660 \\
        CTracker~\cite{peng2020chained}    & 33.4 & 42.8 & 39.4 & 24.5 & 2,987 \\
 
        \textbf{GateMOT (Ours)} & \textbf{48.4} & \textbf{67.8} & \textbf{64.5} & \textbf{44.7} & \textbf{1,058} \\
        \bottomrule
      \end{tabular}%
    }
\end{table}
 
\begin{table}[t]
  \centering
    \caption{Comparison with representative methods on \textbf{SportsMOT}.}
    \label{tab:gate_sportsmot}
    \resizebox{\columnwidth}{!}{%
      \setlength{\tabcolsep}{4pt}
      \begin{tabular}{@{}l @{\hspace{4pt}\vrule\hspace{4pt}} ccccc@{}}
        \toprule
        \textbf{Tracker} & \textbf{HOTA} & \textbf{MOTA} & \textbf{IDF1} & \textbf{AssA} & \textbf{DetA} \\
        \midrule
 
        \textit{Kalman filter:} \\
        FairMOT~\cite{zhang2021fairmot}     & 49.3 & 86.4 & 53.5 & 34.7 & 70.2 \\
        ByteTrack~\cite{zhang2022bytetrack}     & 64.1 & 95.9 & 71.4 & 52.3 & 78.5 \\
        OC\text{-}SORT~\cite{Cao2023Observation} & 73.7 & 96.5 & 74.0 & 61.5 & 88.5 \\
        MixSORT~\cite{Cui2023Sportsmot} & 74.1 & 96.5 & 74.4 & 62.0 & 88.5 \\
        \midrule
 
        \textit{Query\text{-}based:} \\
        TransTrack~\cite{sun2020transtrack}  & 68.9 & 92.6 & 71.5 & 57.5 & 82.7 \\
        GTR~\cite{zhou2022global}             & 54.5 & 67.9 & 55.8 & 45.9 & 64.8 \\
        MeMOTR~\cite{gao2023memotr}       & 70.0 & 91.5 & 71.4 & 59.1 & 83.1 \\
        MOTIP~\cite{gao2025multiple}                  & 72.6 & 92.4 & 77.1 & 63.2 & 83.5 \\
        SambaMOTR~\cite{segu2024samba}      & 70.5  & 90.4 & 73.3  & 60.6  & 82.2 \\
        \midrule
 
        \textit{Learnable motion:} \\
        TrackSSM~\cite{hu2024trackssm}     & 74.4  & 96.8 & 74.5  & 62.4  & 88.8 \\
        DiffMOT~\cite{lv2024diffmot} & 76.2 & \textbf{97.1} & 76.1 & 65.1 & \textbf{89.3} \\
        MambaTrack~\cite{xiao2024mambatrack}    & 72.6  & 95.3 & 72.8  & 60.3  & 87.6 \\
        ETTrack~\cite{han2025ettrack}       & 74.3  & 96.8 & 74.5  & 62.1  & 88.8 \\
        GeneralTrack~\cite{qin2024towards}   & 74.1 & 96.8  & 76.4  & 61.7  & 89.0    \\
 
        \textbf{GateMOT (Ours)} & \textbf{76.3} & 96.5 & \textbf{79.0} & \textbf{65.8} & 88.5 \\
        \bottomrule
      \end{tabular}%
    }
\end{table}
 
\begin{table}[t]
  \centering
    \caption{Comparison with representative methods on \textbf{MOT17}.}
    \label{tab:gate_mot17}
    \resizebox{\columnwidth}{!}{%
      \setlength{\tabcolsep}{4pt}
      \begin{tabular}{@{}l @{\hspace{4pt}\vrule\hspace{4pt}} ccccc@{}}
        \toprule
        \textbf{Tracker} & \textbf{HOTA} & \textbf{MOTA} & \textbf{IDF1} & \textbf{AssA} & \textbf{DetA} \\
        \midrule
 
        \textit{Kalman filter:} \\
        FairMOT~\cite{zhang2021fairmot} & 59.3 & 73.7 & 72.3 & 58.0 & 60.9 \\
        ByteTrack~\cite{zhang2022bytetrack} & 63.1 & 80.3 & 77.3 & 62.0 & 64.5 \\
        OC\text{-}SORT~\cite{Cao2023Observation} & 63.2 & 78.0 & 77.5 & 63.2 & 63.2 \\
        SparseTrack~\cite{liu2025sparsetrack} & 65.1 & 81.0 & 80.1 & 65.1 & 65.3 \\
        TOPICTrack~\cite{Cao2025TOPIC} & 63.9 & 78.8 & 78.7 & 64.3 & -- \\
        FeatureSORT~\cite{hashempoor2025featuresort} & 63.0 & 79.6 & 77.2 & 62.0 & 64.4 \\
        Hybrid-SORT~\cite{yang2024hybrid} & 63.6 & 79.3 & 78.4 & -- & -- \\
        \midrule
 
        \textit{Query\text{-}based:} \\
        TransTrack~\cite{sun2020transtrack}  & 54.1 & 75.2 & 63.5 & 47.9 & 61.6 \\
        TrackFormer~\cite{meinhardt2022trackformer} & 57.3 & 74.1 & 68.0 & 54.1 & 60.9 \\
        MOTR~\cite{zeng2022motr}         & 57.8 & 73.4 & 68.6 & 55.7 & 60.3 \\
        MOTRv2~\cite{zhang2023motrv2}       & 62.0 & 78.6 & 75.0 & 60.6 & 63.8 \\
        MeMOTR~\cite{gao2023memotr}       & 58.8 & 72.8 & 71.5 & 58.4 & 59.6 \\
        MeMOT~\cite{Cai2022MeMOT}         & 56.9  & 72.5  & 69.0  & 55.2  & --    \\
        \midrule
 
        \textit{Learnable motion:} \\
        TrackSSM~\cite{hu2024trackssm}     & 61.4  & 78.5  & 74.1  & 59.6  & \textbf{63.6} \\
        MambaTrack~\cite{xiao2024mambatrack}   & 61.1  & 78.1  & 73.9  & --    & --    \\
        STDFormer~\cite{hu2023stdformer}       & 60.9  & 78.4  & 73.1  & 58.4  & --    \\
        ETTrack~\cite{han2025ettrack}         & 61.9  & \textbf{79.0}  & 75.9  & 60.5  & --    \\
        DiffusionTrack~\cite{luo2024diffusiontrack}         & 60.8  & 77.9  & 73.8  & 58.8  & 63.2    \\
 
        \textbf{GateMOT (Ours)} & \textbf{63.3} & 78.0 & \textbf{77.9} & \textbf{63.5} & 63.4 \\
        \bottomrule
      \end{tabular}%
    }
\end{table}
 
\begin{table}[t]
  \centering
    \caption{Comparison with representative methods on \textbf{MOT20}.}
    \label{tab:gate_mot20}
    \resizebox{\columnwidth}{!}{%
      \setlength{\tabcolsep}{4pt}
      \begin{tabular}{@{}l @{\hspace{4pt}\vrule\hspace{4pt}} ccccc@{}}
        \toprule
        \textbf{Tracker} & \textbf{HOTA} & \textbf{MOTA} & \textbf{IDF1} & \textbf{AssA} & \textbf{DetA} \\
        \midrule
 
        \textit{Kalman filter:} \\
        FairMOT~\cite{zhang2021fairmot} & 54.6 & 61.8 & 67.3 & 54.7 & 54.7 \\
        ByteTrack~\cite{zhang2022bytetrack} & 61.3 & 77.8 & 75.2 & 59.6 & 63.4 \\
        BoT\text{-}SORT~\cite{Aharon2022BoTSORT} & 63.3 & 77.8 & 77.5 & 62.9 & 64.0 \\
        OC\text{-}SORT~\cite{Cao2023Observation} & 62.1 & 75.5 & 75.9 & 62.0 & -- \\
        SparseTrack~\cite{liu2025sparsetrack} & 63.5 & 78.1 & 77.6 & 63.1 & 64.1\\
        TOPICTrack~\cite{Cao2025TOPIC} & 62.6 & 72.4 & 77.6 & 65.4 & -- \\
        FeatureSORT~\cite{hashempoor2025featuresort} & 61.3 & 76.6 & 75.1 & 60.1 & 62.7 \\
        Hybrid-SORT~\cite{yang2024hybrid} & 62.5 & 76.4 & 76.2 & -- & -- \\
        \midrule
 
        \textit{Query\text{-}based:} \\
        MeMOT~\cite{Cai2022MeMOT} & 54.1 & 63.7 & 66.1 & 55.0 & -- \\
        MOTRv2~\cite{zhang2023motrv2} & 60.3 & 76.2 & 72.2 & 58.1 & 62.9 \\
        \midrule
 
        \textit{Learnable motion:} \\
        STDFormer~\cite{hu2023stdformer} & 60.1 & 65.2 & 70.1 & -- & -- \\
        DiffusionTrack~\cite{luo2024diffusiontrack} & 55.3 & 72.8 & 66.3 & 51.3 & 59.9 \\
        MotionTrack~\cite{qin2023motiontrack} & \textbf{62.8} & \textbf{78.0} & 76.5 & 61.8 & \textbf{64.0} \\
        OFTrack~\cite{song2025temporal} & 61.9 & 75.3 & 74.7 & 62.1 & 62.9 \\
        DiffMOT~\cite{lv2024diffmot} & 61.7 & 76.7 & 74.9 & 60.5 & 63.2 \\
 
        \textbf{GateMOT (Ours)} & \textbf{62.8} & 77.6 & \textbf{77.3} & \textbf{64.1} & 61.9 \\
        \bottomrule
      \end{tabular}%
    }
\end{table}
 
\paragraph{Results on SportsMOT.}
SportsMOT contains long sequences of fast-moving athletes with similar uniforms and body types. GateMOT achieves the best HOTA (76.3), AssA (65.8), and IDF1 (79.0) among the compared methods (Table~\ref{tab:gate_sportsmot}), supporting effective identity discrimination by task-specific Q-Attention.
 
\paragraph{Results on MOT17.}
MOT17 combines camera motion, viewpoint variation, and frequent pedestrian occlusion in complex urban scenes. GateMOT achieves 63.3 HOTA, 78.0 MOTA, 77.9 IDF1, 63.5 AssA, and 63.4 DetA (Table~\ref{tab:gate_mot17}). These balanced scores indicate reliable localization and association.
\paragraph{Results on MOT20.}
MOT20 focuses on extremely crowded scenes with heavy occlusion and dense interactions. Under the official protocol, GateMOT achieves 62.8 HOTA, 77.6 MOTA, 77.3 IDF1, 64.1 AssA, and 61.9 DetA (Table~\ref{tab:gate_mot20}), demonstrating robust tracking under severe crowding.
 
\subsection{Ablation Studies}
\label{sec:ablation}
 
To dissect GateMOT, we evaluate its core components, compare Q-Attention with representative decoder alternatives, and analyze efficiency and transferability. Unless specified otherwise, all ablation models use a DLA-34 backbone and a one-layer decoder under identical training settings, with experiments conducted on BEE24-test and MOT17-val.
 
\begin{figure}[t]
\centering
\setlength{\tabcolsep}{0pt}
\renewcommand{\arraystretch}{0}
\begin{tabular}{@{}cccc@{}}
\stackinset{l}{2pt}{t}{2pt}{\colorbox{black}{\textcolor{white}{\tiny\bfseries Q-Attn}}}
    {\stackinset{r}{5pt}{b}{5pt}{\textcolor{yellow}{\scriptsize\bfseries \#10}}
    {\includegraphics[width=.25\linewidth]{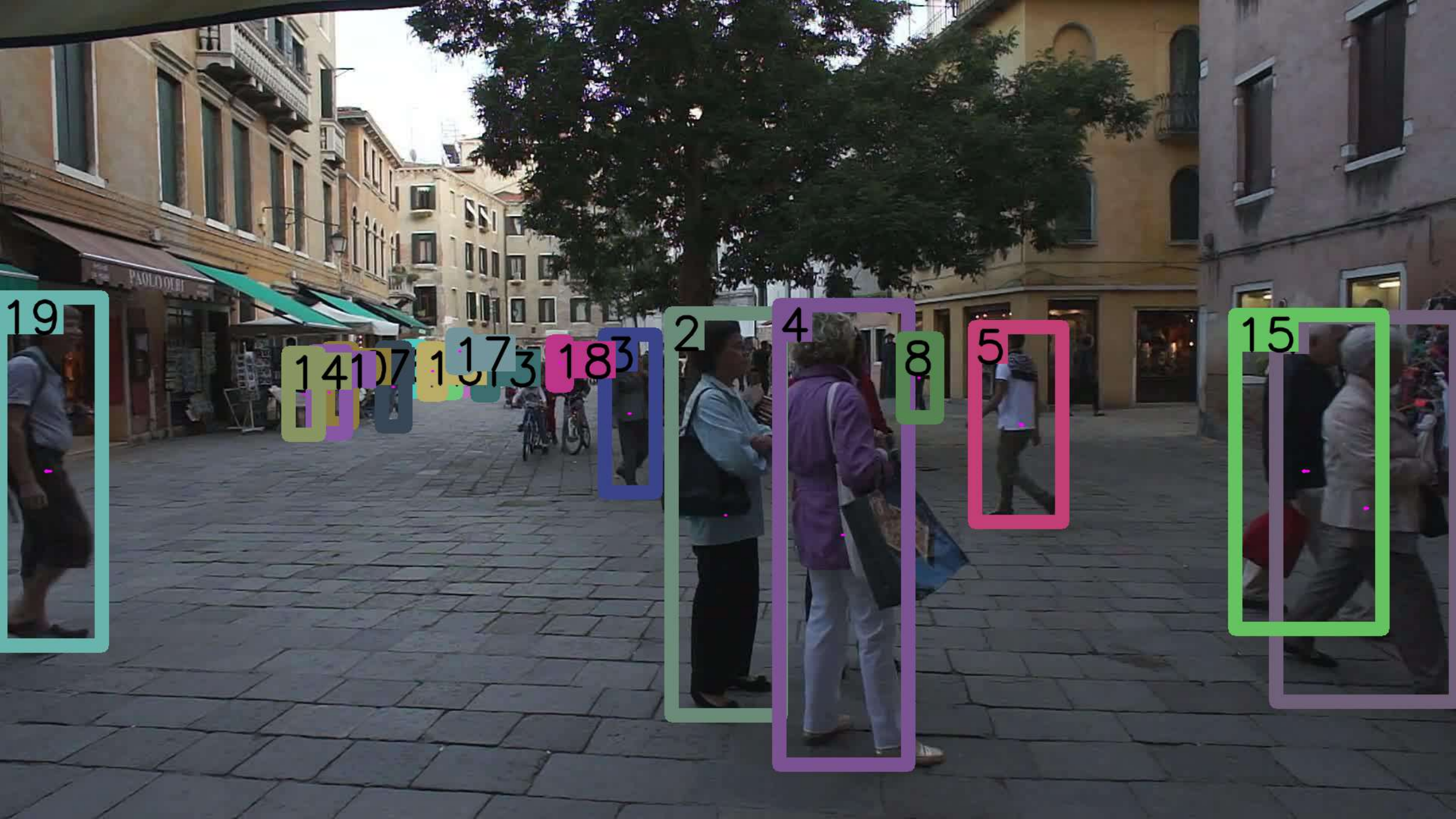}}} &
\stackinset{r}{5pt}{b}{5pt}{\textcolor{yellow}{\scriptsize\bfseries \#120}}
    {\includegraphics[width=.25\linewidth]{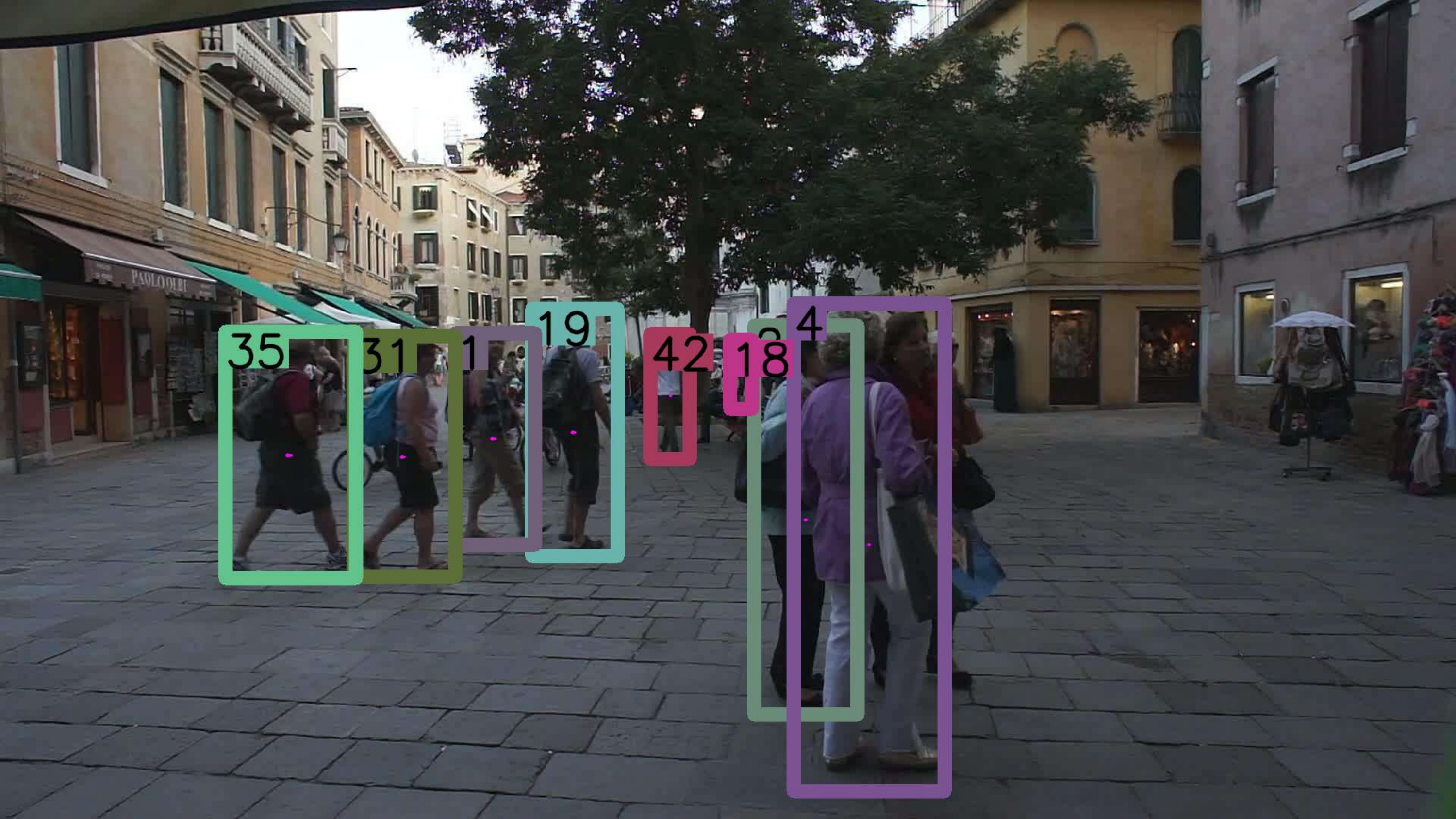}} &
\stackinset{r}{5pt}{b}{5pt}{\textcolor{yellow}{\scriptsize\bfseries \#170}}
    {\includegraphics[width=.25\linewidth]{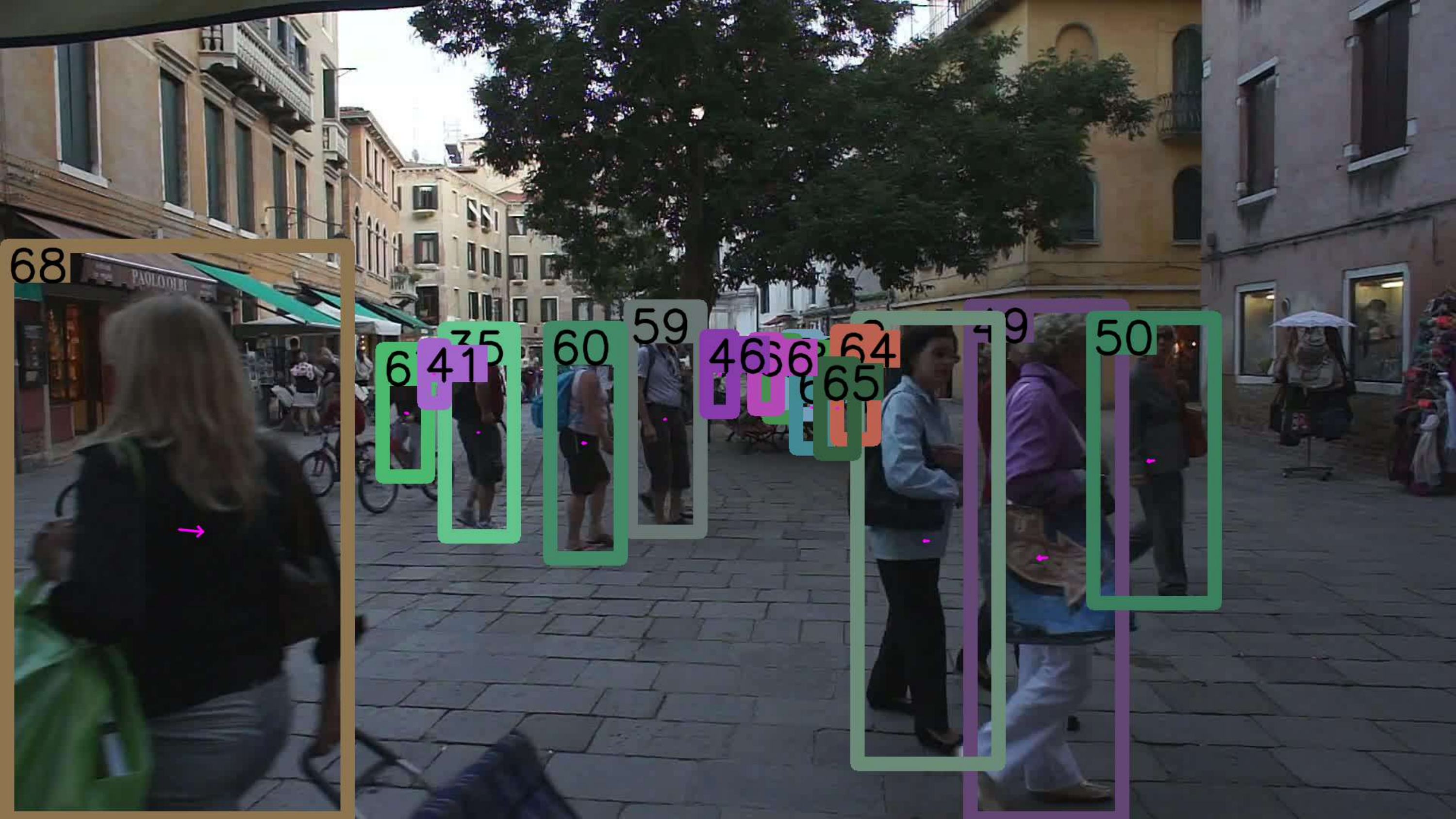}} &
\stackinset{r}{5pt}{b}{5pt}{\textcolor{yellow}{\scriptsize\bfseries \#270}}
    {\includegraphics[width=.25\linewidth]{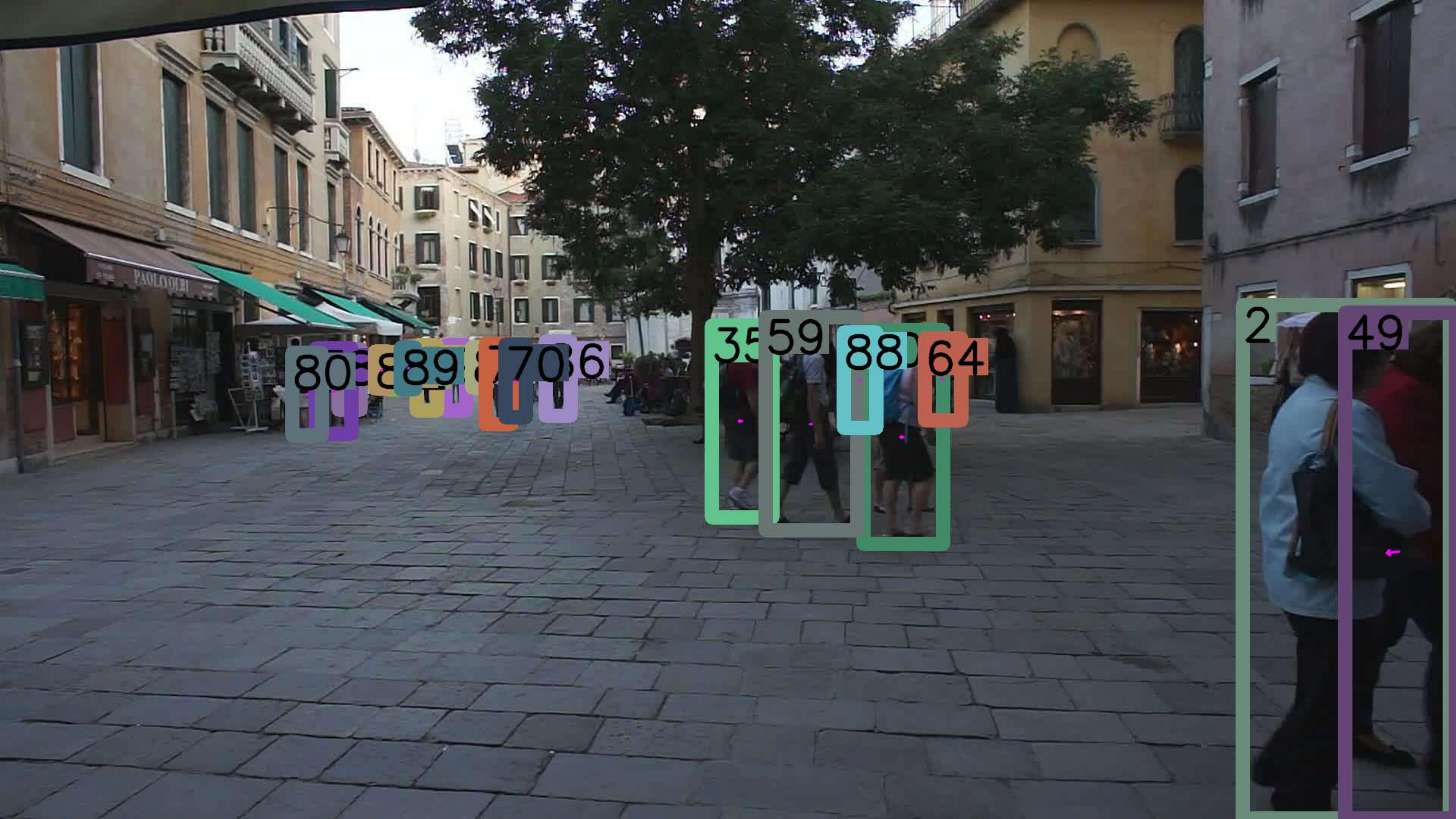}} \\
\stackinset{l}{2pt}{t}{2pt}{\colorbox{black}{\textcolor{white}{\tiny\bfseries Conv}}}
    {\stackinset{r}{5pt}{b}{5pt}{\textcolor{yellow}{\scriptsize\bfseries \#10}}
    {\includegraphics[width=.25\linewidth]{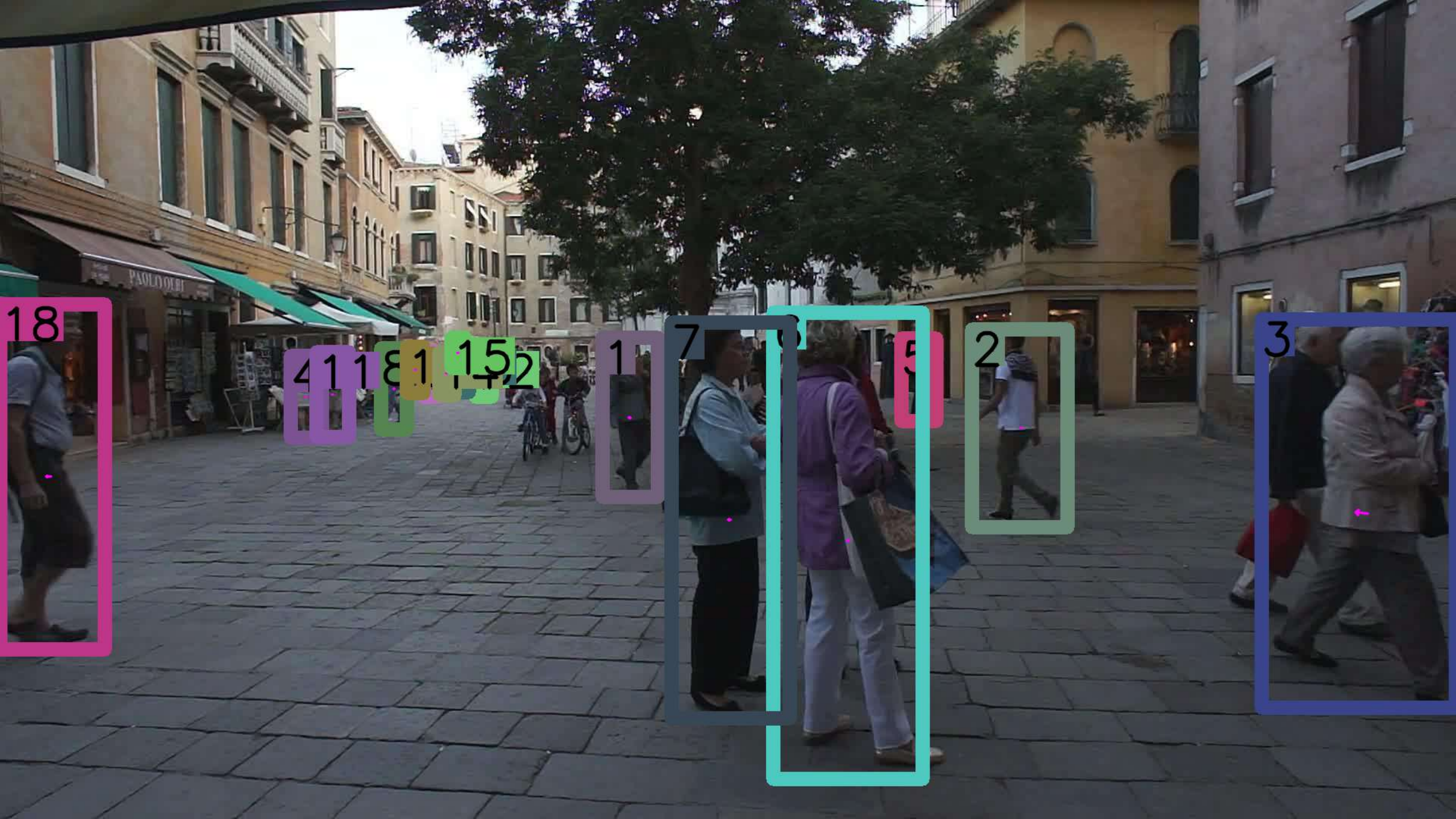}}} &
\stackinset{r}{5pt}{b}{5pt}{\textcolor{yellow}{\scriptsize\bfseries \#120}}
    {\includegraphics[width=.25\linewidth]{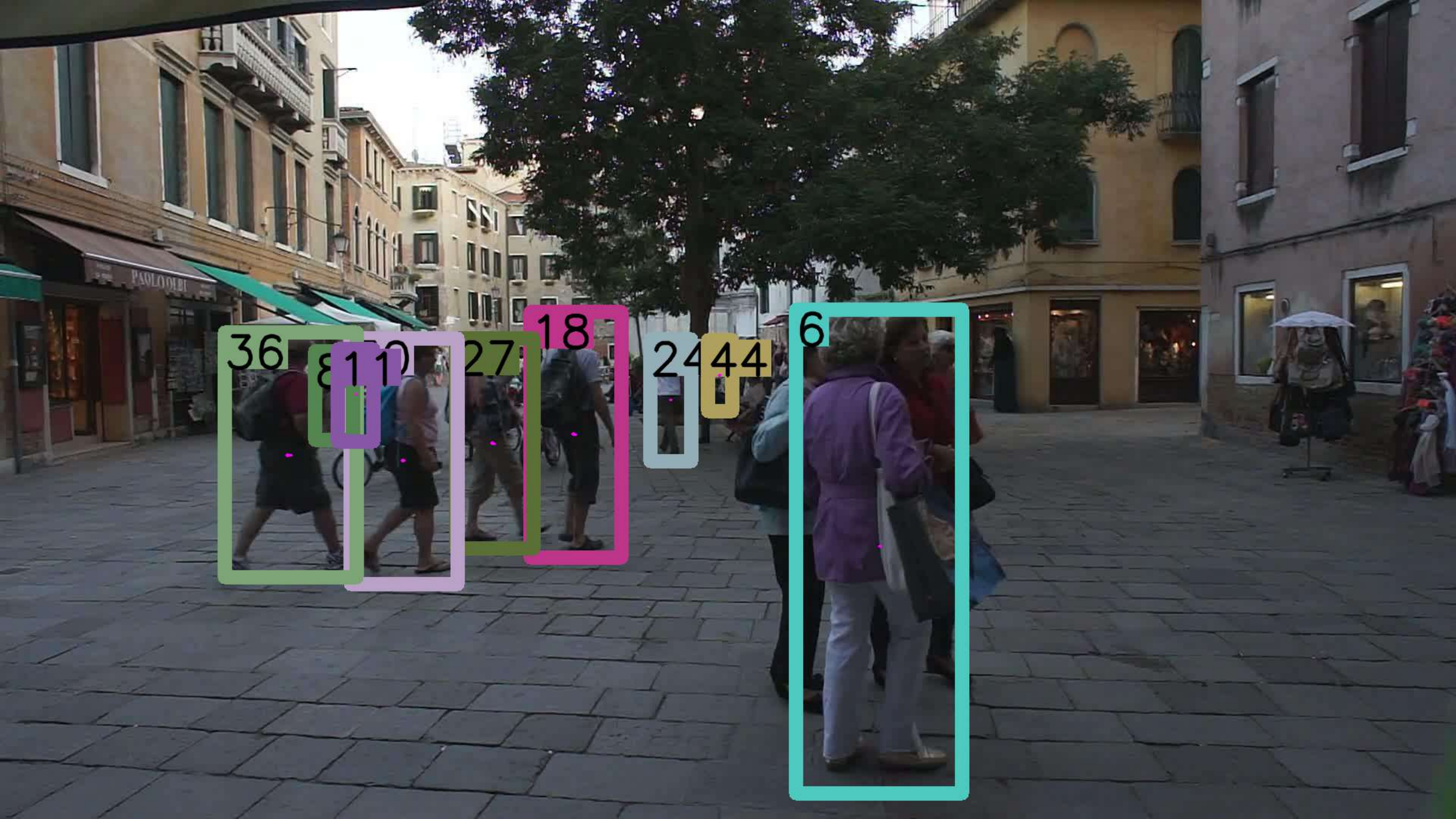}} &
\stackinset{r}{5pt}{b}{5pt}{\textcolor{yellow}{\scriptsize\bfseries \#170}}
    {\includegraphics[width=.25\linewidth]{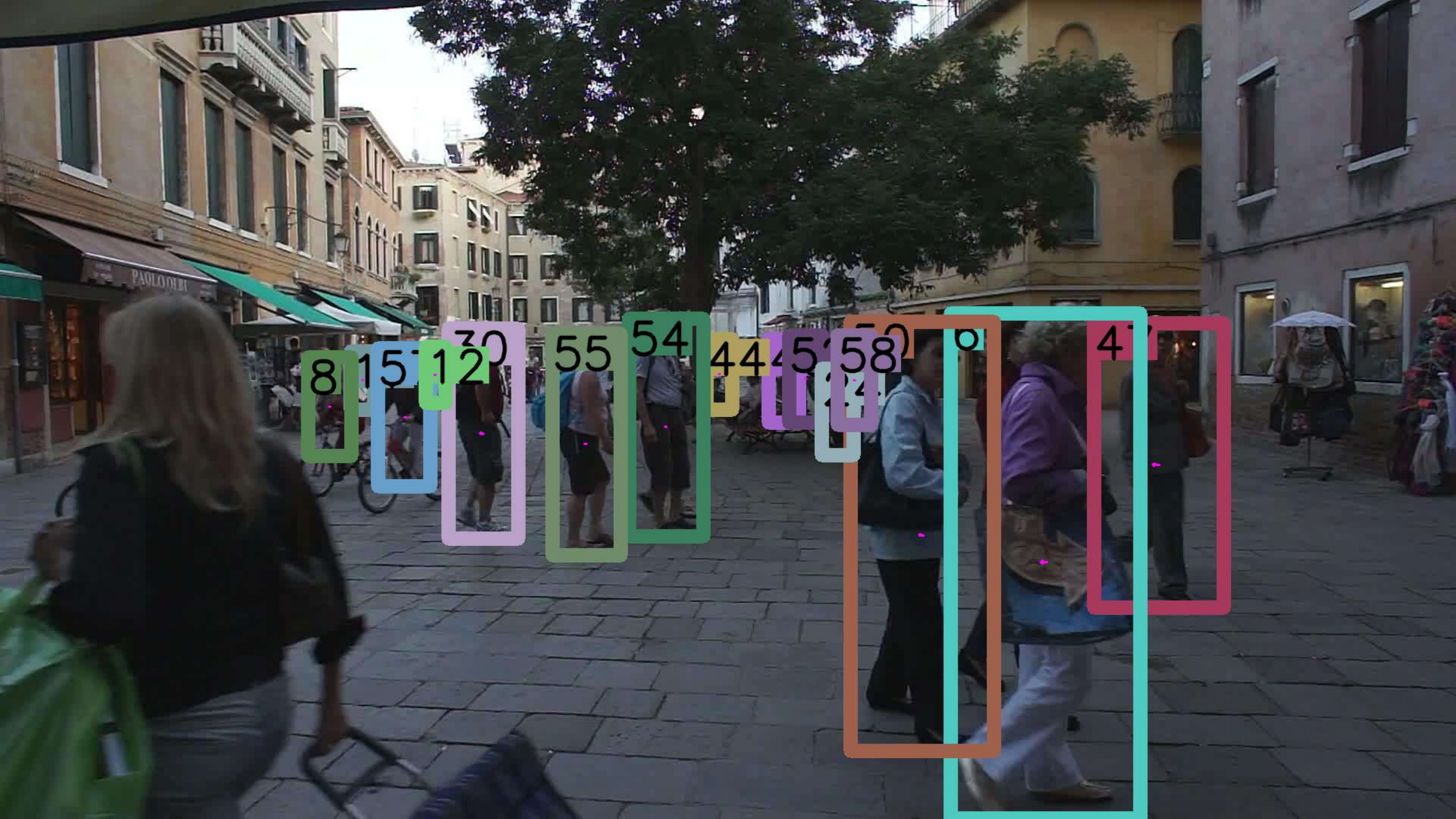}} &
\stackinset{r}{5pt}{b}{5pt}{\textcolor{yellow}{\scriptsize\bfseries \#270}}
    {\includegraphics[width=.25\linewidth]{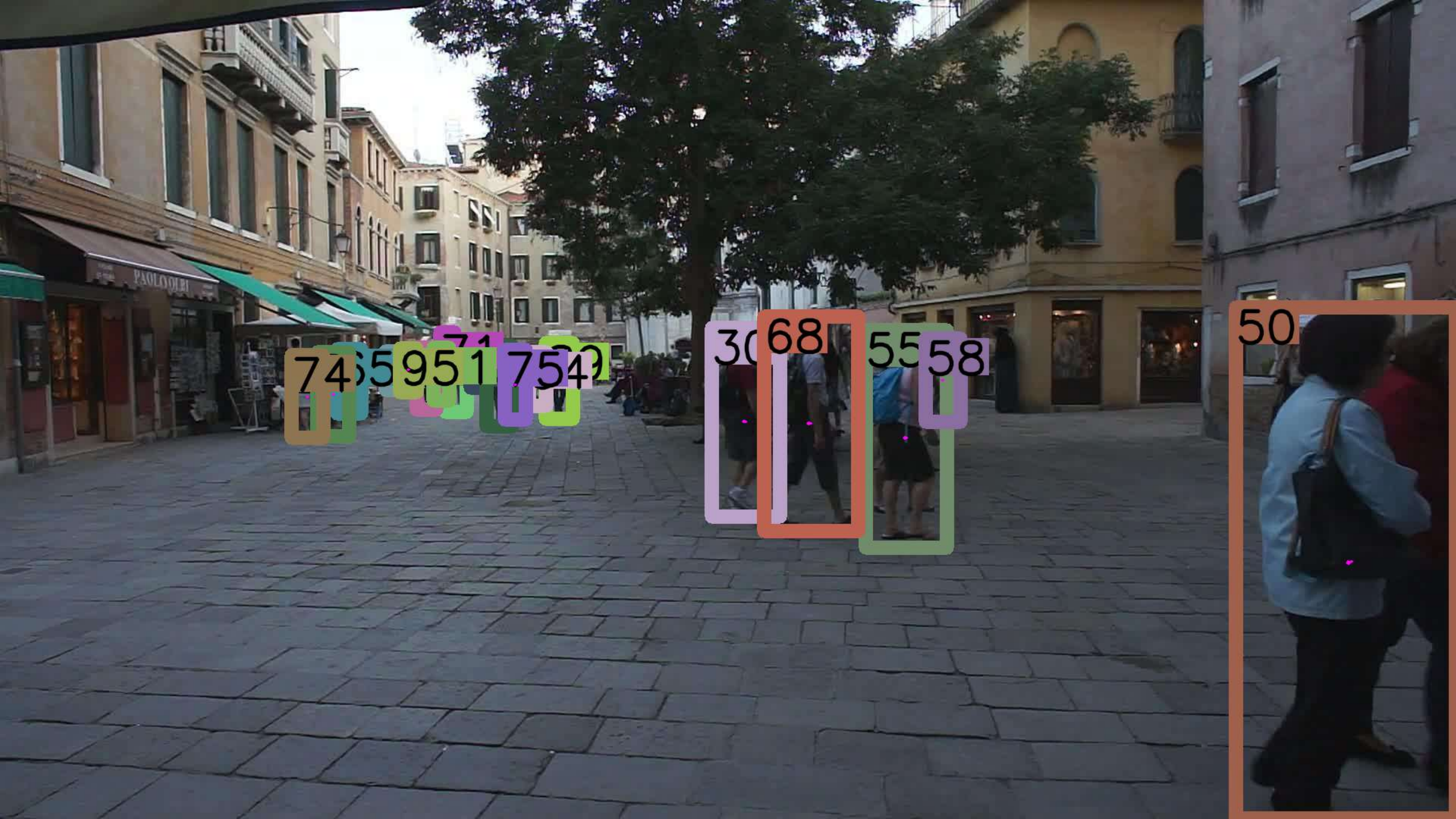}} \\
\stackinset{l}{2pt}{t}{2pt}{\colorbox{black}{\textcolor{white}{\tiny\bfseries Self-Attn}}}
    {\stackinset{r}{5pt}{b}{5pt}{\textcolor{yellow}{\scriptsize\bfseries \#10}}
    {\includegraphics[width=.25\linewidth]{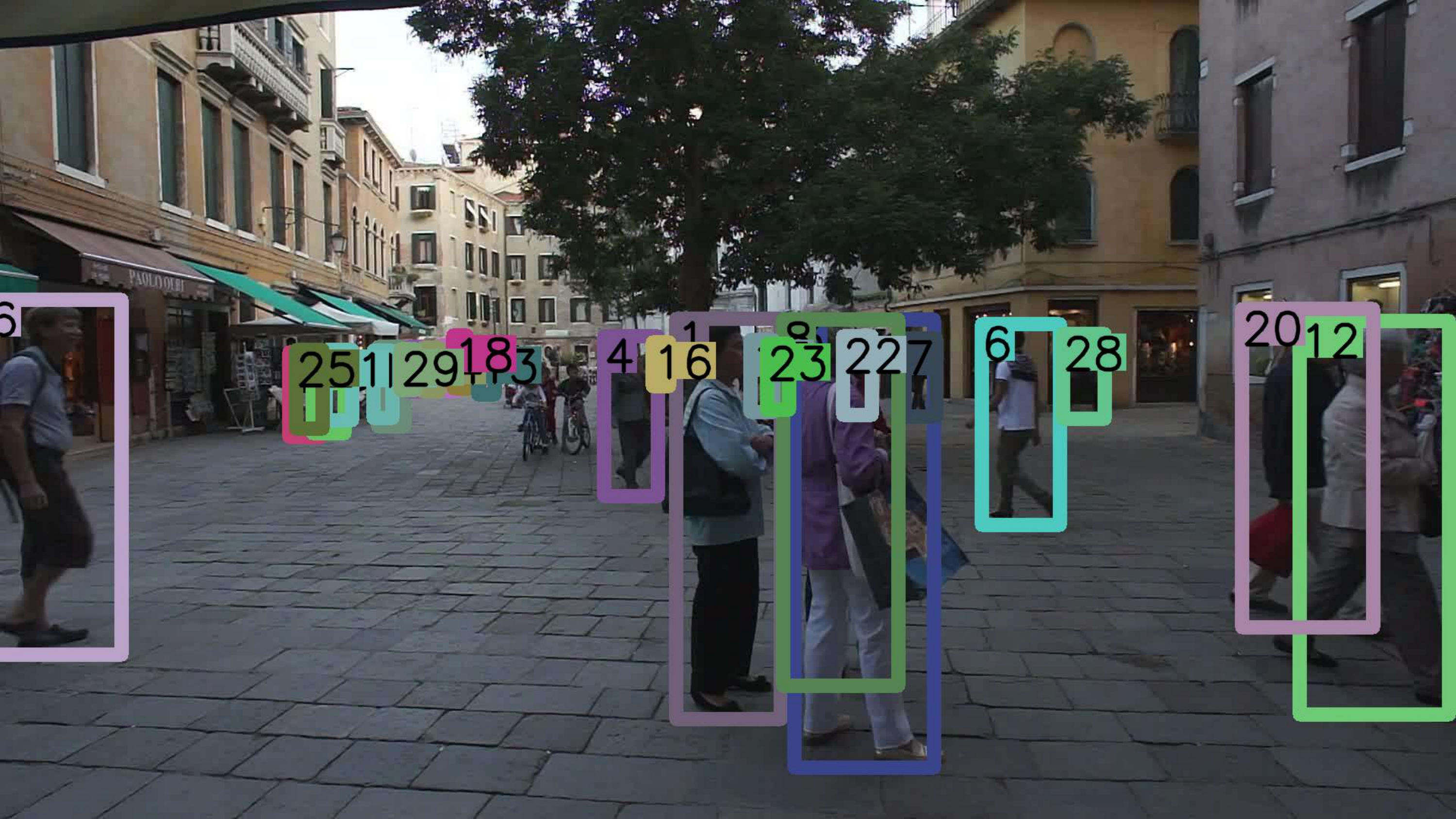}}} &
\stackinset{r}{5pt}{b}{5pt}{\textcolor{yellow}{\scriptsize\bfseries \#120}}
    {\includegraphics[width=.25\linewidth]{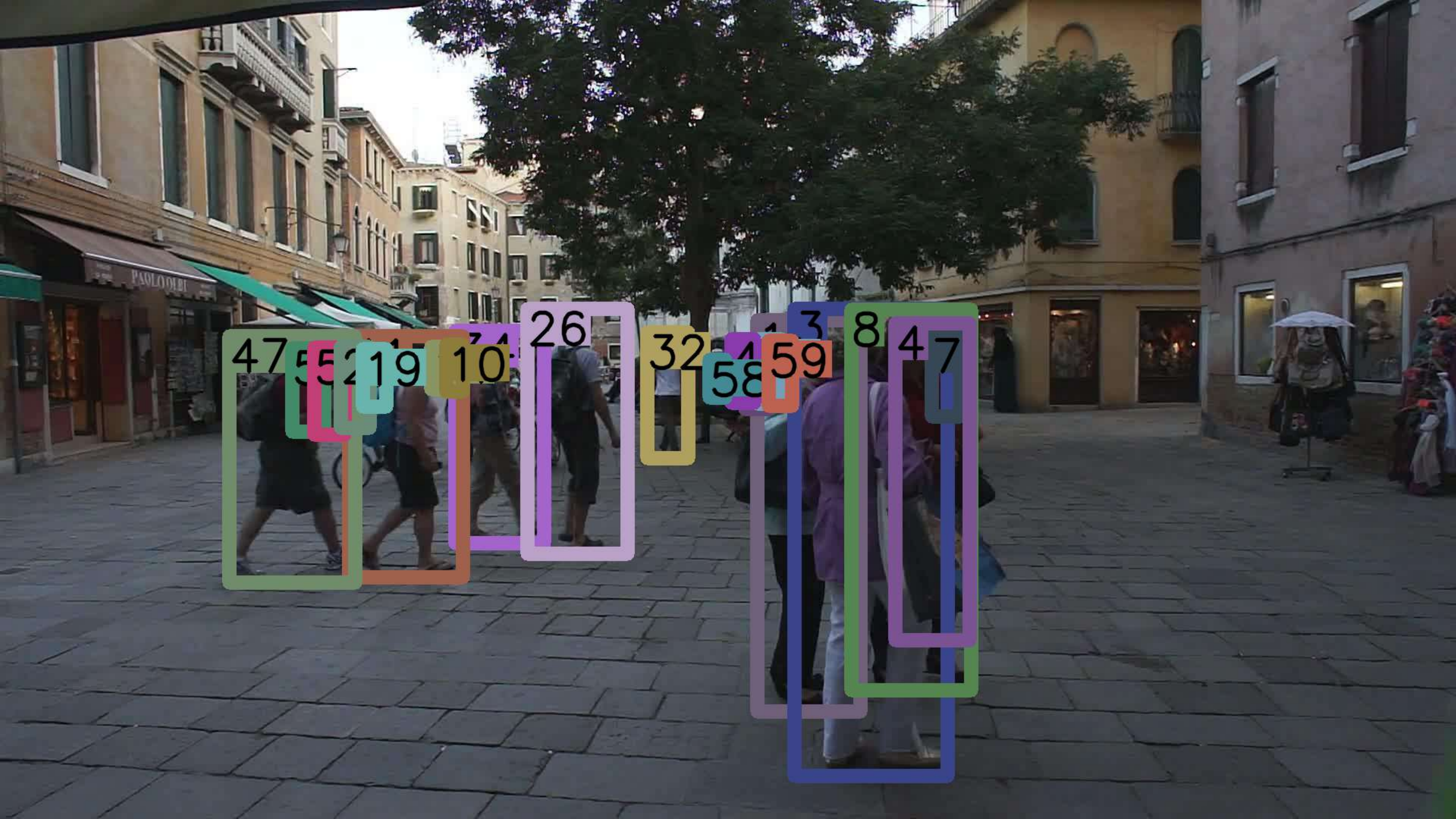}} &
\stackinset{r}{5pt}{b}{5pt}{\textcolor{yellow}{\scriptsize\bfseries \#170}}
    {\includegraphics[width=.25\linewidth]{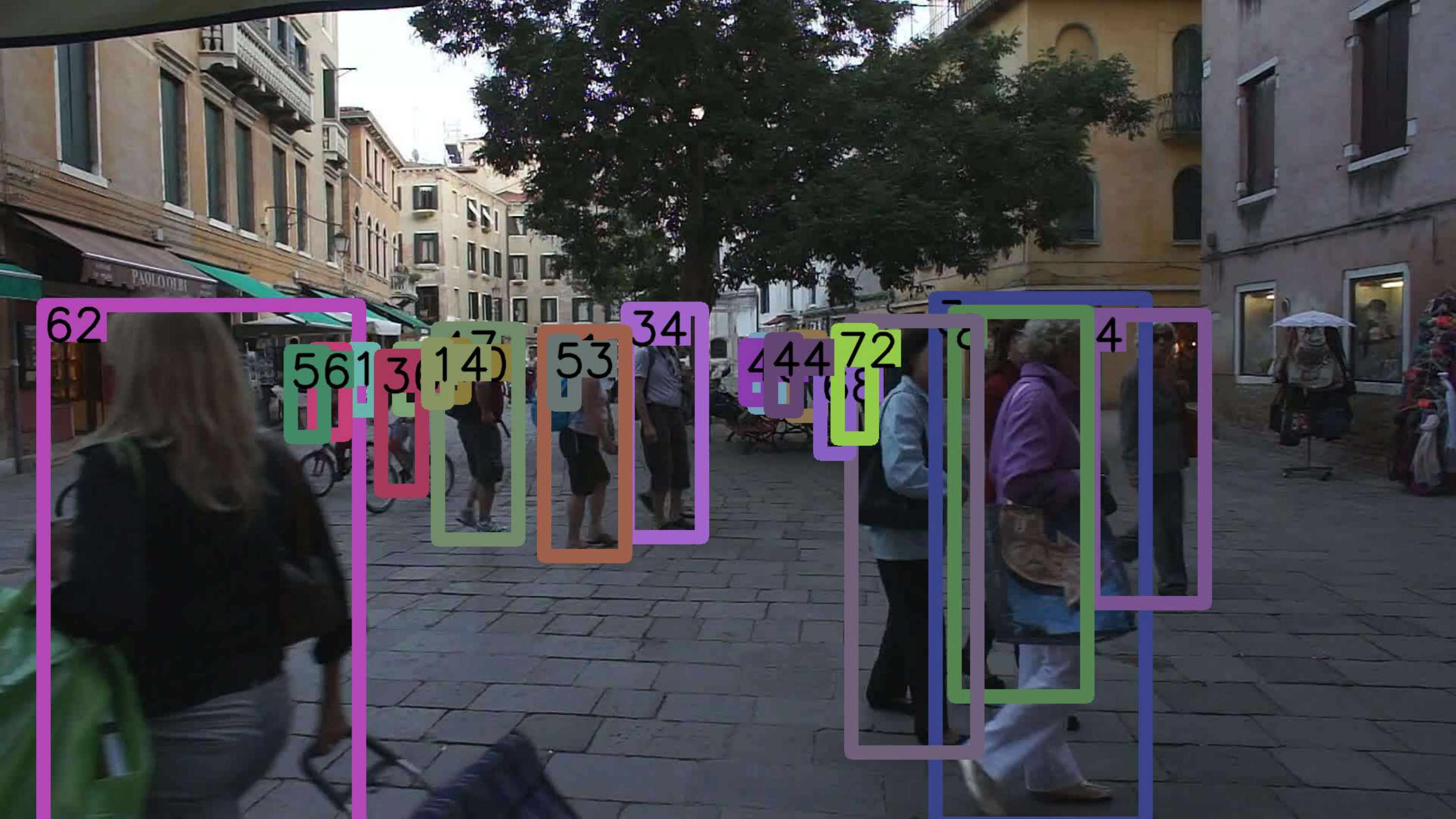}} &
\stackinset{r}{5pt}{b}{5pt}{\textcolor{yellow}{\scriptsize\bfseries \#270}}
    {\includegraphics[width=.25\linewidth]{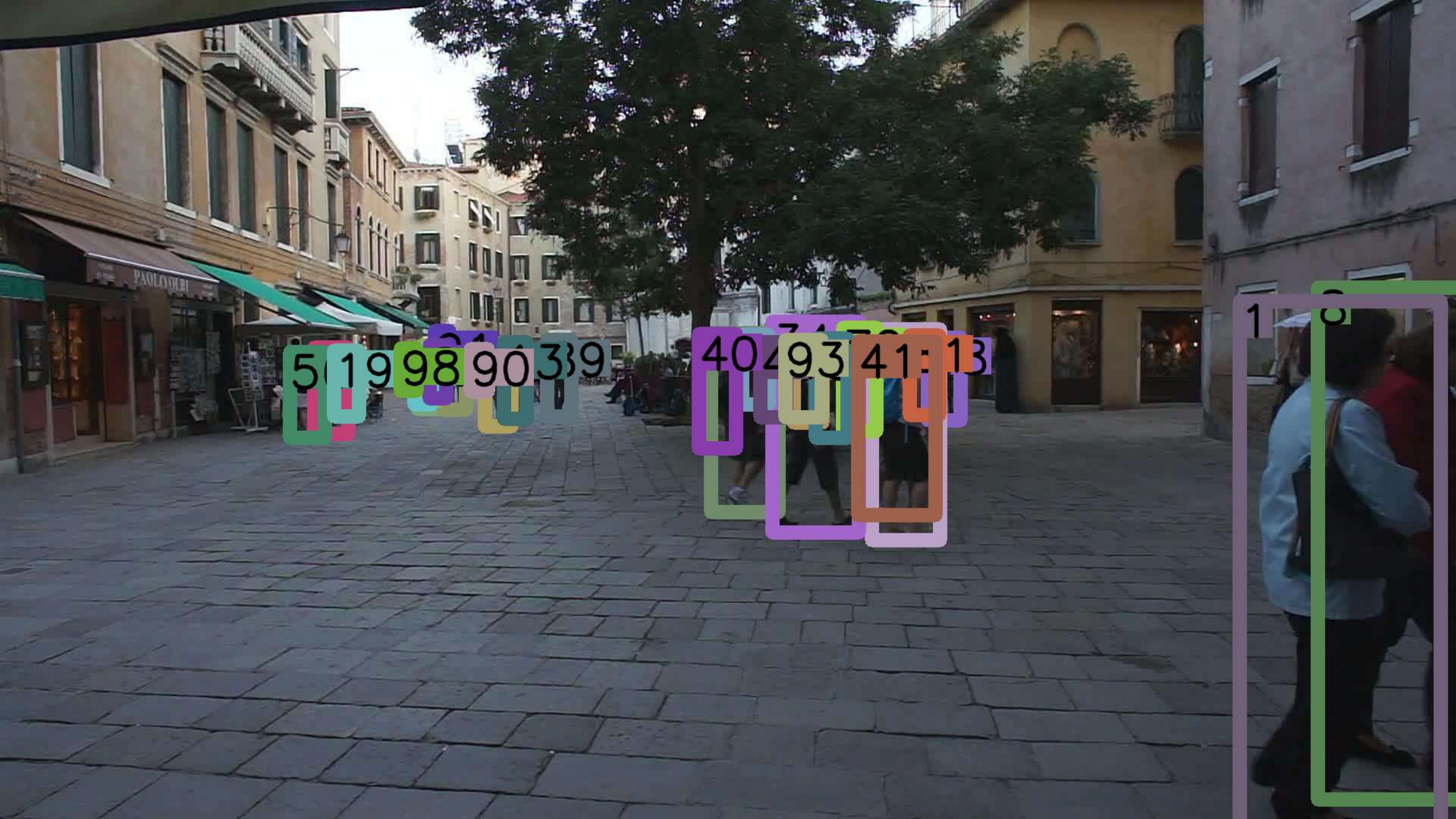}}
\end{tabular}
\caption{\textbf{Qualitative comparison of decoder architectures on MOT17-val (depth=1).} Rows show Q-Attention, the convolutional baseline, and vanilla self-attention at the same frames. Q-Attention produces more consistent tracking outputs under occlusion and crowding.}
\label{fig:decoder_vis}
\end{figure}

\paragraph{Analysis of Decoder Architectures.}
Table~\ref{tab:ablation_decoder} compares Q-Attention with stacked convolutional heads and vanilla self-attention. A single Q-Attention layer outperforms both a three-layer Conv head and a one-layer self-attention head, while stacking two or three Q-Attention layers produces small declines on both datasets, indicating that one lightweight layer is sufficient. The slight degradation may result from repeated gating and local aggregation over-smoothing localization-sensitive features, since dense tracking depends on fine-grained instance separability. We do not evaluate deeper self-attention variants because their computation and memory grow rapidly with the number of spatial locations; at the high resolutions required for dense tracking, a two-layer head already exceeds available hardware. Figure~\ref{fig:decoder_vis} further compares the three designs on the same MOT17-val sequence and shows that Q-Attention maintains more consistent tracking outputs under crowding and occlusion.
 
\paragraph{Analysis of Efficiency, Scalability, and Transferability.}
Table~\ref{tab:ablation_efficiency} evaluates decoder efficiency and backbone scaling on MOT17-val. With DLA-34, Q-Attention achieves 64.4 HOTA at 159.7 GFLOPs and 13.3 FPS, outperforming vanilla self-attention, deformable attention, and linear attention in both accuracy and efficiency. Scaling the DLA backbone from DLA-34 to DLA-60 and DLA-169 increases HOTA from 64.4 to 65.8 and 66.7, with the expected cost--speed trade-off. ResNet-50~\cite{koonce2021resnet} and YOLOX-X~\cite{ge2021yolox} retain 65.2 and 64.9 HOTA with practical throughput, demonstrating transferability across backbone families.
 
\paragraph{Analysis of Existing Gating Mechanisms.}
Table~\ref{tab:main_gating_variants} compares Q-Attention with representative recalibration and gating modules under identical training settings. Q-Attention further improves both tracking and association metrics through prediction-head-specific channel-spatial gating, local Key consolidation, and residual Value preservation.
 
\begin{table}[t]
  \centering
    \caption{Performance comparison of different decoder architectures and depths.}
    \label{tab:ablation_decoder}
    \resizebox{\columnwidth}{!}{%
      \setlength{\tabcolsep}{4pt}
      \begin{tabular}{@{}lc @{\hspace{4pt}\vrule\hspace{4pt}} ccc @{\hspace{4pt}\vrule\hspace{4pt}} ccc@{}}
        \toprule
        \multicolumn{2}{c}{\textbf{Decoder Head}} &
        \multicolumn{3}{c}{\textbf{BEE24-test}} &
        \multicolumn{3}{c}{\textbf{MOT17-val}} \\
        \cmidrule(lr){1-2}\cmidrule(lr){3-5}\cmidrule(lr){6-8}
        \textbf{Type} & \textbf{Layers} &
        \textbf{HOTA} & \textbf{MOTA} & \textbf{IDF1} &
        \textbf{HOTA} & \textbf{MOTA} & \textbf{IDF1} \\
        \midrule
        \multirow{3}{*}{Conv} & 1 & 44.3 & 65.4 & 59.3 & 60.2 & 76.6 & 75.5 \\
        & 2 & 45.1 & 65.9 & 60.1 & 62.1 & 78.8 & 75.7 \\
        & 3 & 45.9 & 66.0 & 60.4 & 62.2 & 77.3 & 76.8 \\
        \midrule
        \multirow{2}{*}{Attn} & 1 & 46.3 & 65.8 & 59.8 & 63.0 & 78.7 & 76.6 \\
        & 2 & \multicolumn{6}{c}{Exceeds Memory Limit} \\
        \midrule
        \rowcolor{gray!25}
        \multirow{3}{*}{\textbf{Q-Attn}} & \textbf{1} & \textbf{48.4} & \textbf{67.8} & \textbf{64.5} & \textbf{64.4} & \textbf{79.5} & \textbf{77.5} \\
        & 2 & 47.9 & 67.4 & 64.1 & 64.1 & 78.4 & 77.4 \\
        & 3 & 48.0 & 66.9 & 62.8 & 64.2 & 79.3 & 77.3 \\
        \bottomrule
      \end{tabular}%
    }
\end{table}
 
\begin{table}[t]
  \centering
    \caption{Analysis of computational efficiency and model scalability on MOT17-val (one-layer decoder).}
    \label{tab:ablation_efficiency}
    \resizebox{\columnwidth}{!}{%
      \setlength{\tabcolsep}{4pt}
      \begin{tabular}{@{}ll @{\hspace{4pt}\vrule\hspace{4pt}} cccc@{}}
        \toprule
        \textbf{Model / Head} & \textbf{Backbone} & \textbf{Params (M)}$\downarrow$ & \textbf{GFLOPs}$\downarrow$ & \textbf{FPS} & \textbf{HOTA} \\
        \midrule
        Conv Head & \multirow{5}{*}{DLA-34} & 21.85 & 157.3 & 14.8 & 60.2 \\
        Self-Attn Head & & 22.30 & 268.4 & 4.6 & 63.0 \\
        Deformable Attn & & 22.40 & 182.5 & 11.8 & 63.8 \\
        Linear Attn & & 22.30 & 168.3 & 12.5 & 63.5 \\
        \rowcolor{gray!25}
        \textbf{Q-Attn Head} & & \textbf{22.05} & \textbf{159.7} & \textbf{13.3} & \textbf{64.4} \\
        \midrule
        \multirow{4}{*}{Q-Attn Head} & DLA-60 & 38.92 & 290.1 & 11.0 & 65.8 \\
        & DLA-169 & 70.64 & 468.7 & 6.8 & 66.7 \\
        & ResNet-50 & 24.30 & 175.3 & 11.4 & 65.2 \\
        & YOLOX-X & 27.90 & 198.6 & 10.8 & 64.9 \\
        \bottomrule
      \end{tabular}%
    }
\end{table}
 
\begin{table}[t]
\centering
\caption{Comparison with representative gating mechanisms under controlled settings.}
\label{tab:main_gating_variants}
\resizebox{\columnwidth}{!}{%
\begin{tabular}{@{}lccc@{}}
\toprule
\textbf{Mechanism} & \textbf{BEE24-test HOTA} & \textbf{BEE24-test AssA} & \textbf{MOT17-val HOTA} \\
\midrule
Conv (no gating) & 44.3 & 38.4 & 60.2 \\
SE~\cite{hu2018squeeze} & 46.1 & 39.2 & 62.8 \\
CBAM-Spatial~\cite{woo2018cbam} & 46.8 & 40.5 & 63.4 \\
CBAM-Full & 47.3 & 41.1 & 63.7 \\
\rowcolor{gray!25}
\textbf{Q-Attention} & \textbf{48.4} & \textbf{44.7} & \textbf{64.4} \\
\bottomrule
\end{tabular}}
\end{table}
 
\paragraph{Analysis of Resolution Scaling.}
Table~\ref{tab:ablation_resolution_scaling} compares Conv, Q-Attn, and Self-Attn at three resolutions under the same depth=1 setting. Q-Attn stays close to Conv in GFLOPs/memory/FPS across all scales, while Self-Attn grows sharply in cost and latency.
 
\paragraph{Analysis of Key Q-Attention Components.}
Table~\ref{tab:ablation_qattn_components} evaluates the selection--preservation architecture. Although gating alone improves HOTA over the Conv baseline, its MOT17-val IDF1 decreases from 75.5 to 75.3. Compared with the gated-Key-only variant, the full module improves HOTA by 2.2 points on BEE24-test and 1.8 points on MOT17-val, with corresponding IDF1 gains of 2.7 and 2.2 points. These gains support preserving unfiltered Value information alongside selective gating. Removing local aggregation reduces HOTA by 1.7 and 1.5 points, respectively, indicating its complementary role in consolidating local evidence. Replacing learnable fusion with addition or using Tanh instead of Sigmoid also degrades performance.
 
\begin{table}[t]
  \centering
    \caption{Efficiency scaling with input resolution under depth=1 decoder.}
    \label{tab:ablation_resolution_scaling}
    \resizebox{\columnwidth}{!}{%
      \setlength{\tabcolsep}{4pt}
      \begin{tabular}{@{}ll @{\hspace{4pt}\vrule\hspace{4pt}} cccc@{}}
        \toprule
        \textbf{Head} & \textbf{Resolution} & \textbf{Params (M)}$\downarrow$ & \textbf{GFLOPs}$\downarrow$ & \textbf{Mem (GB)}$\downarrow$ & \textbf{FPS} \\
        \midrule
        Conv Head & & 21.85 & 157.3 & 2.14 & 14.8 \\
        \rowcolor{gray!25}
        \textbf{Q-Attn Head} & & \textbf{22.05} & \textbf{159.7} & \textbf{2.27} & \textbf{13.3} \\
        Self-Attn Head & \multirow{-3}{*}{$608 \times 1088$} & 22.30 & 268.4 & 11.25 & 4.6 \\
        \midrule
 
        Conv Head & & 21.85 & 262.8 & 3.46 & 10.4 \\
        \rowcolor{gray!25}
        \textbf{Q-Attn Head} & & \textbf{22.05} & \textbf{265.9} & \textbf{3.79} & \textbf{9.6} \\
        Self-Attn Head & \multirow{-3}{*}{$800 \times 1440$} & 22.30 & 754.8 & 21.37 & 1.8 \\
        \midrule
 
        Conv Head & & 21.85 & 473.1 & 6.28 & 6.3 \\
        \rowcolor{gray!25}
        \textbf{Q-Attn Head} & & \textbf{22.05} & \textbf{478.6} & \textbf{6.76} & \textbf{5.9} \\
        Self-Attn Head & \multirow{-3}{*}{$1080 \times 1920$} & 22.30 & 2687.5 & 39.84 & 0.7 \\
        \bottomrule
      \end{tabular}%
    }
\end{table}
 
\begin{table}[!t]
  \centering
    \caption{Ablation on key Q-Attention components.}
    \label{tab:ablation_qattn_components}
    \resizebox{\columnwidth}{!}{%
      \setlength{\tabcolsep}{4pt}
      \begin{tabular}{@{}l @{\hspace{4pt}\vrule\hspace{4pt}} cc @{\hspace{4pt}\vrule\hspace{4pt}} cc@{}}
        \toprule
        \multirow{2}{*}{\textbf{Variant}} & \multicolumn{2}{c}{\textbf{BEE24-test}} & \multicolumn{2}{c}{\textbf{MOT17-val}} \\
        \cmidrule(lr){2-3} \cmidrule(lr){4-5}
        & \textbf{HOTA} & \textbf{IDF1} & \textbf{HOTA} & \textbf{IDF1} \\
        \midrule
        Baseline (Conv head) & 44.3 & 59.3 & 60.2 & 75.5 \\
        Gated Key only (w/o $\mathbf{V}$) & 46.2 & 61.8 & 62.6 & 75.3 \\
        w/o local aggregation & 46.7 & 62.4 & 62.9 & 75.8 \\
        Addition fusion & 47.1 & 62.9 & 63.4 & 76.2 \\
        Tanh gate & 47.5 & 63.3 & 63.7 & 76.6 \\
        \rowcolor{gray!25}
        \textbf{Full Q-Attn (Ours)} & \textbf{48.4} & \textbf{64.5} & \textbf{64.4} & \textbf{77.5} \\
        \bottomrule
      \end{tabular}%
    }
\end{table}
 
\section{Conclusion}
\label{sec:conclusion}

We presented GateMOT, a dense object tracker built on Q-Gated Attention, which repurposes Query into a spatially adaptive gate that selects task-relevant Key evidence while an ungated Value pathway preserves complementary information. By removing quadratic pairwise interactions, Q-Attention equips parallel detection, motion, and re-identification heads with task-specific feature selection at near-convolutional cost. Experiments on four benchmarks show a strong accuracy--efficiency trade-off, the best BEE24-test results among compared methods, and consistent gains across backbones, establishing Q-Attention as a practical building block for dense tracking.

\ifdefined\includesupplement
\clearpage
\input{sec/7_suppl}
\fi

\bibliography{aaai2027}

\end{document}